\documentclass{article}

\usepackage{amsmath}
\usepackage{arxiv}

\usepackage[utf8]{inputenc} 
\usepackage[T1]{fontenc}    
\usepackage{hyperref}       
\usepackage{url}            
\usepackage{booktabs}       
\usepackage{amsfonts}       
\usepackage{nicefrac}       
\usepackage{microtype}      
\usepackage{lipsum}		
\usepackage{graphicx}
\usepackage{natbib}
\usepackage{doi}

\date{}
\title{HingeRLC-GAN: Combating Mode Collapse  with Hinge Loss and RLC Regularization}

\author{ {Osman Goni}\\
	Department of Computer Science and Engineering\\
	United International  University\\
	Dhaka, Bangladesh \\
	\texttt{ogoni202046@bscse.uiu.ac.bd} \\
	\And
	{Himadri Saha Arka} \\
	Department of Computer Science and Engineering\\
	United International University\\
	Dhaka, Bangladesh \\
	\texttt{harka202008@bscse.uiu.ac.bd} \\
    \And
	{Mithun Halder} \\
	Department of Computer Science and Engineering\\
	United International University\\
	Dhaka, Bangladesh \\
	\texttt{mhalder201041@bscse.uiu.ac.bd}
    \And
	{Mir Moynuddin Ahmed
Shibly} \\
	Department of Computer Science and Engineering\\
	United International University\\
	Dhaka, Bangladesh \\
	\texttt{moynuddin@cse.uiu.ac.bd}
    \And
	{Swakkhar Shatabda} \\
	Department of Computer Science and Engineering\\
	BRAC University\\
	Dhaka, Bangladesh \\
	\texttt{swakkhar.shatabda@bracu.ac.bd}
}

\renewcommand{\shorttitle}{Combating Mode Collapse  with Hinge Loss and RLC Regularization}

\hypersetup{
pdftitle={A template for the arxiv style},
pdfsubject={q-bio.NC, q-bio.QM},
pdfauthor={David S.~Hippocampus, Elias D.~Striatum},
pdfkeywords={First keyword, Second keyword, More},
}

\begin{document}
\maketitle

\begin{abstract}
Recent advances in Generative Adversarial Networks (GANs) have demonstrated their capability for producing high-quality images. However, a significant challenge remains mode collapse, which occurs when the generator produces a limited number of data patterns that do not reflect the diversity of the training dataset. This study addresses this issue by proposing a number of architectural changes aimed at increasing the diversity and stability of GAN models. We start by improving the loss function with Wasserstein loss and Gradient Penalty to better capture the full range of data variations. We also investigate various network architectures and conclude that ResNet significantly contributes to increased diversity. Building on these findings, we introduce HingeRLC-GAN, a novel approach that combines RLC Regularization and the Hinge loss function. With a FID Score of 18 and a KID Score of 0.001, our approach outperforms existing methods by effectively balancing training stability and increased diversity.
\end{abstract}

\keywords{GAN  \and Diversity \and Mode Collapse \and Hinge Loss \and Regularization \and ResNet \and Fréchet inception distance (FID) \and Kernel Inception Distance (KID)}

\begin{figure*}[!htb]
\centering
\includegraphics[width=1\textwidth]{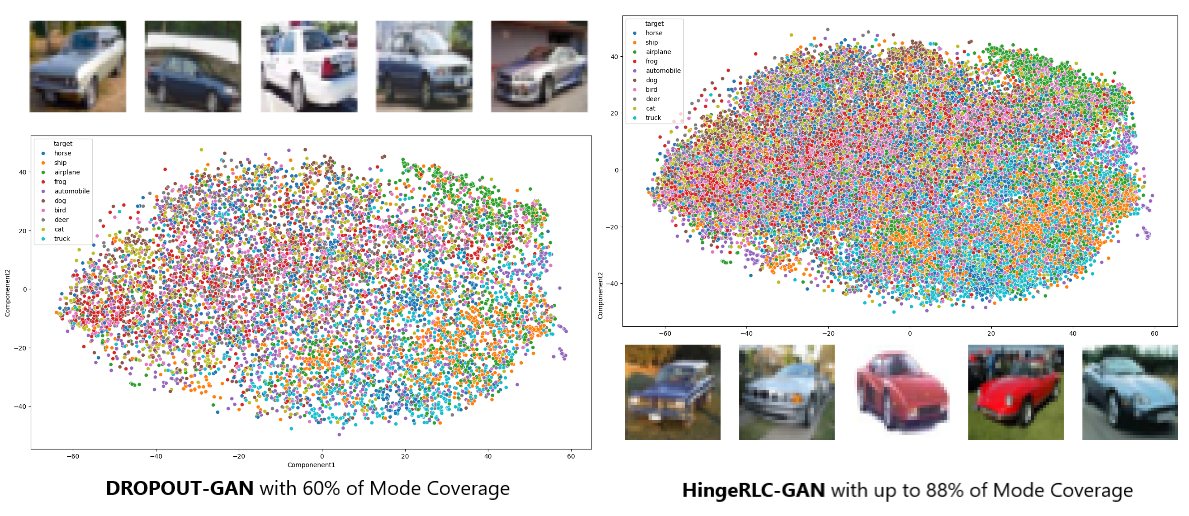}
\caption{Mode Coverage: DROPOUT-GAN (left),HingeRLC-GAN (right). The proposed method is performing up to 30\% better in mode capture.}
\label{figModeCoverage}
\end{figure*}

\section{Introduction}

Generative Adversarial Networks (GANs) \cite{goodfellow2014generative} have made remarkable strides in generating high-fidelity images. These models are foundational for many vision applications, including data augmentation \cite{shorten2019survey,antoniou2017data}, domain adaptation \cite{tzeng2017adversarial,hoffman2018cycada}, image extrapolation \cite{van2021image}, image-to-image translation \cite{isola2017image,zhu2017unpaired}, and image editing \cite{shen2020interpreting,shen2020interfacegan}. However, the success of GANs often hinges on the availability of large, diverse training datasets, which can be costly and labor-intensive to compile \cite{robinson2019lapgan}.

A significant challenge associated with GANs is mode collapse, where the generator produces a limited variety of outputs and fails to capture the full diversity of the data distribution \cite{arjovsky2017wasserstein,lin2021wgan}. This issue drastically reduces the diversity of generated data, limiting the utility of GANs across different applications. Mode collapse is exemplified in Figure~\ref{figModeCoverage}, which demonstrates how a generator may inadequately represent the modes of the data distribution.

To address the issue of mode collapse in GANs, extensive research has explored a variety of new techniques. These include advanced loss functions, such as Wasserstein loss with Gradient Penalty \cite{gulrajani2017improved,isola2017image}, which have shown promise in stabilizing GAN training and enhancing output diversity by overcoming the limitations of traditional loss functions. Research into multi-generator models, like MAD GAN \cite{ghosh2018multi}, aims to improve diversity by utilizing multiple generators in conjunction with a single discriminator. Additionally, innovative methods, such as employing orthogonal vectors to address mode collapse \cite{li2021tackling}, have further advanced our understanding of GANs and their training challenges.

In this paper, we focus on improving GAN performance to enhance mode coverage, particularly for small datasets. Our approach consists of several stages:

\begin{itemize}
\item First, we evaluated various GAN architectures to identify the most effective structures for generating diverse outputs.
\item Next, we examined different loss functions and regularization techniques to find the optimal combination for mitigating mode collapse.
\item Finally, we developed a model that integrates RLC regularization \cite{tseng2021regularizing} with Hinge Loss \cite{lim2017geometric}, and performed a comprehensive comparison of our HingeRLC-GAN against conventional GAN models.
\end{itemize}

Our research aims to determine the most effective architectural components, loss functions, and regularization techniques to improve mode coverage and produce high-quality, diverse synthetic images, especially when working with small datasets.

\section{Related Work}
\label{sec:lit-review}
Arnab Ghosh et al. introduced MAD-GAN, a model that leverages multiple generators and a single discriminator to improve sample diversity \cite{ghosh2018multi}. In this setup, the discriminator not only distinguishes real from fake samples but also identifies which generator produced each fake sample, promoting a wider range of generated outputs.

Wei Li et al. developed a method employing orthogonal vectors to address mode collapse in multi-generator frameworks \cite{li2021tackling}. Their approach involves extracting feature vectors from generator outputs and minimizing their orthogonality to preserve diversity, using a new minimax formula to enhance convergence and balance.

Jae Hyun Lim and Jong Chul Ye proposed Geometric GAN, which redefines adversarial training through geometric steps involving hyperplane separation to overcome issues such as vanishing gradients and instability \cite{lim2017geometric}. This SVM-inspired method improves the reliability and efficiency of training.

Mordido et al. introduced Dropout-GAN, which applies dropout regularization to a discriminator ensemble to combat overfitting and maintain diversity in generated samples \cite{mordido2018dropout}. Dropout-GAN demonstrates superior performance compared to other variants by generating diverse and realistic data while minimizing the Fréchet distance.

Sen Pei et al. presented the Pluggable Diversity Penalty Module (PDPM), which enforces diversity in the feature space using normalized Gram matrices \cite{pei2021alleviating}. PDPM achieves outstanding results across various tasks, surpassing traditional methods such as ALI, DCGAN, and MSGAN.

Pan et al. introduced UniGAN, which aims to address u-mode collapse by focusing on uniform diversity \cite{pan2022unigan}. This model employs a generator based on Normalizing Flow and a regularization technique to ensure uniform output diversity, allowing for seamless integration with other frameworks.

\section{Proposed Method}
\label{sec:Methodology}

First, we examine the GAN architecture, focusing on its core components and overall structure. Next, we review the loss functions used during GAN training, emphasizing their roles and how they influence the model's performance. We then delve into regularization techniques, assessing their impact on stabilizing training and improving the model's generalization capabilities. Finally, we explore the effectiveness of the HingeRLC-GAN by investigating its architectural modifications and their contributions to enhancing diversity and training stability.

\subsection{Architectural Overview}

Generative Adversarial Networks (GANs) are a class of machine learning frameworks designed for generating realistic data. A GAN comprises two neural networks: the Generator \( G \) and the Discriminator \( D \). These networks are trained simultaneously in a competitive framework. The Generator creates synthetic data with the goal of approximating real data, while the Discriminator's task is to distinguish between real and generated data. The GAN framework involves training the Generator \( G \) and Discriminator \( D \) through a minimax game, which is formalized as follows:

\[
\min_G \max_D \mathbb{E}_{x \sim p_{\text{data}}(x)} [\log D(x)] + \mathbb{E}_{z \sim p_z(z)} [\log (1 - D(G(z)))]
\]

In this formulation, the Generator \( G \) aims to minimize the likelihood of the Discriminator correctly identifying generated data as fake, while the Discriminator \( D \) seeks to maximize its ability to differentiate between real and generated data. Here, \( p_{\text{data}}(x) \) represents the distribution of real data, and \( p_z(z) \) represents the distribution of the input noise vector \( z \). We have experimented with various architectures such as DenseNet, MobileNet, and EfficientNet, and found that the ResNet architecture consistently produces superior results.

\subsubsection{Generator}

The Generator architecture is constructed using ResNet blocks, which incorporate residual connections to support effective gradient flow. The detailed architecture is as follows:

\[
\text{Noise} \; z \xrightarrow{\text{Linear}(128 \rightarrow 4 \times 4 \times 128)} \xrightarrow{\text{Reshape}(-1, 128, 4, 4)} \xrightarrow{\text{ResNet Block}_1} \xrightarrow{\text{ResNet Block}_2} 
\]
\[
\xrightarrow{\text{ResNet Block}_3} \xrightarrow{\text{BatchNorm2d}} \xrightarrow{\text{ReLU}} \xrightarrow{\text{Conv2d}(3, 3, \text{padding}='same')} \xrightarrow{\text{Tanh}} \xrightarrow{\text{Output Image}}
\]

Each ResNet Block consists of:

\[
\text{CCBN}(128, 10) \rightarrow \text{ReLU} \rightarrow \text{Upsample}
\]
\[
(\text{scale factor}=2) \rightarrow \text{Conv2d}(128, 3, \text{padding}='same')
\]

Here, \(\text{CCBN}(128, 10)\) denotes Conditional Batch Normalization with 128 channels and 10 conditions. The upsampling layer increases the spatial dimensions of the feature maps, and the convolutional layer with a kernel size of 3 and `same' padding ensures that the output maintains the required dimensions.

\begin{figure}[h!]
    \centering
    \includegraphics[width=0.8\textwidth]{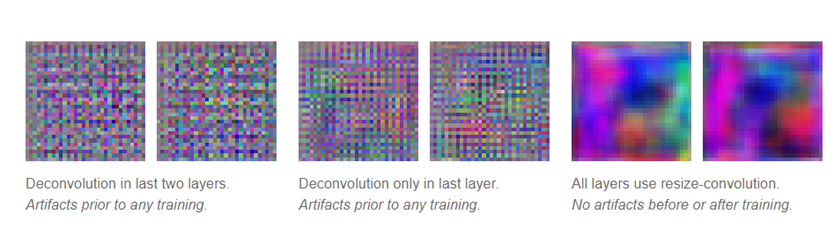}
    \caption{Artifact produced by `Conv2dTranspose' layers with checkerboard patterns.}
    \label{fig:artifact}
\end{figure}
Avoiding the use of `Conv2dTranspose` layers is crucial for minimizing artifacts in generated images. `Conv2dTranspose' layers \cite{odena2016deconvolution} with a stride of 2 are commonly used for upsampling, effectively doubling the size of the images. However, they can introduce artifacts such as checkerboard patterns due to uneven overlapping of the layers. This problem arises from the inherent characteristics of the `Conv2dTranspose' operation rather than from adversarial training itself.

To address this issue, our architecture utilizes `Upsample' layers followed by `Conv2d' layers for upsampling. This approach avoids the artifacts typically associated with `Conv2dTranspose' layers. Figure \ref{fig:artifact} illustrates artifacts produced by a generator using `Conv2dTranspose' in the last two layers. By employing `Upsample' layers combined with `Conv2d' layers, we achieve cleaner, artifact-free images and improved overall image quality.

\subsubsection{Discriminator}

The Discriminator uses ResNet blocks with downsampling to classify images. Its architecture is as follows:

\[
\text{Image} \; x \xrightarrow{\text{Concatenate}(\text{Embedding} \; y \; \text{to} \; 32 \times 32)} \xrightarrow{\text{Conv2d}(128, 3, \text{padding}='same')} \xrightarrow{\text{ReLU}} 
\]
\[
\xrightarrow{\text{Conv2d}(128, 3, \text{padding}='same')} 
\xrightarrow{\text{AvgPool2d}(2, 2)} \xrightarrow{\text{ResNet Block Down}_1} \xrightarrow{\text{ResNet Block Down}_2} 
\]
\[
\xrightarrow{\text{ResNet Block Down}_3} \xrightarrow{\text{AdaptiveMaxPool2d}} \xrightarrow{\text{Flatten}} \xrightarrow{\text{Linear}(1)} \xrightarrow{\text{Output Score}}
\]

Each ResNet Block Down consists of:

\[
\text{CCBN}(128, 10) \rightarrow \text{ReLU} \rightarrow \text{Conv2d}(128, 3, \text{padding}='same') \rightarrow \text{AvgPool2d}(2, 2)
\]

Here, \(\text{CCBN}(128, 10)\) denotes Conditional Batch Normalization with 128 channels and 10 conditions. The `AvgPool2d' layer performs downsampling by a factor of 2, reducing the spatial dimensions of the feature maps at each ResNet block.

\begin{figure*}[ht]
    \centering
    \includegraphics[width=1\textwidth]{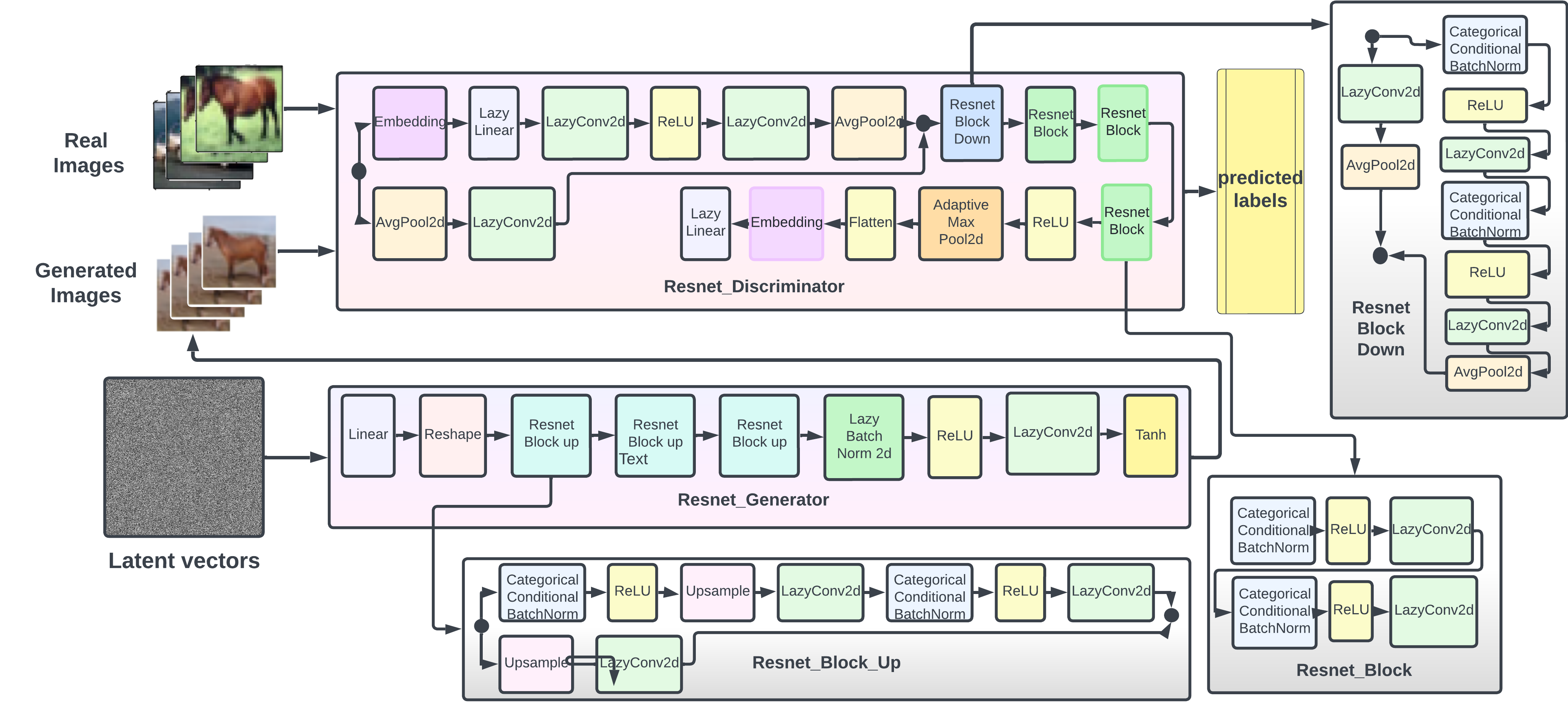}
    \caption{\textbf{HingeRLC-GAN Architecture}: An illustrative example of the ResNetRLC GAN's internal generator and discriminator workings}
    \label{fig:modelarch}
\end{figure*}

The superiority of the ResNet architecture is attributed to several key components:

\begin{enumerate}
    \item \textbf{Residual Connections}: Residual connections in ResNet blocks, defined as:

    \[
    \mathbf{y} = \mathcal{F}(\mathbf{x}, \{W_i\}) + \mathbf{x}
    \]

    allow gradients to flow directly through the network, mitigating the vanishing gradient problem. This is crucial for training deeper networks, as it ensures that gradient updates from the loss function propagate effectively through many layers.

    \item \textbf{Categorical Conditional Batch Normalization (CCBN)}: CCBN \cite{de2017modulating} conditions the normalization process on class labels, enabling class-specific feature generation. The normalization for each feature map \( i \) in class \( c \) is:

    \[
    \text{BN}(x_i, \gamma_{c,i}, \beta_{c,i}) = \gamma_{c,i} \frac{x_i - \mu_i}{\sqrt{\sigma_i^2 + \epsilon}} + \beta_{c,i}
    \]

    where \( \gamma \) and \( \beta \) are learned parameters specific to each class, \( \mu_i \) and \( \sigma_i \) are the mean and variance of the feature maps, and \( \epsilon \) is a small constant for numerical stability.

    \item \textbf{Spectral Normalization}: To enforce Lipschitz continuity, spectral normalization \cite{miyato2018spectral} is applied to the weights of each layer, constraining the largest singular value of the weight matrix \( W \). This is defined as:

    \[
    \frac{W}{\sigma(W)}
    \]

    where \( \sigma(W) \) is the largest singular value of \( W \). Spectral normalization stabilizes the training of the Discriminator, enhancing robustness to variations in the input space.

    \item \textbf{Regularization}: The architecture includes various regularization techniques, such as dropout layers, weight decay, and batch normalization, to prevent overfitting and improve generalization.
\end{enumerate}

Overall, the integration of these elements results in a more stable training process and the generation of higher-quality images compared to other architectures. The ResNet-based GAN leverages deep residual learning, effective class conditioning, and robust normalization techniques to outperform models like DenseNet, MobileNet, and EfficientNet.

\subsection{Loss Functions}
We have experimented with several loss functions, including Binary Cross-Entropy (BCE), Wasserstein Loss, and Least Squares Loss. Our findings indicate that Hinge Loss consistently delivers superior results for our HingeRLC-GAN.

Hinge Loss is defined as:

\[
\mathcal{L}_{\text{Hinge}} = \max(0, 1 - D(x)) + \max(0, 1 + D(G(z)))
\]

This loss function enhances the stability of GAN training by ensuring continuous gradients, even for samples that are correctly classified. This characteristic helps mitigate the vanishing gradient problem, leading to more stable and effective training.

\subsection{Regularization}

In HingeRLC-GAN, we utilize several regularization techniques to enhance training stability and generalization:

\begin{enumerate}
    \item \textbf{Noise:} Introducing noise to the inputs helps prevent the model from overfitting to the training data, promoting better generalization.

\item \textbf{Class Rebalancing:} This technique ensures that the model learns equally from all classes, thereby improving its performance across different categories.

\item \textbf{Gradient Penalty:} By encouraging smoothness in the Discriminator's decision boundary, this technique enhances the model's robustness and stability.
\end{enumerate}

The primary regularization technique employed is Regularized Loss Control (RLC), defined as:

\[
\mathcal{L}_{\text{RLC}} = \mathcal{L}_{\text{Hinge}} + \lambda \sum_{i} \left( \frac{\partial \mathcal{L}}{\partial \theta_i} \right)^2
\]

Here, \( \lambda \) is a hyperparameter that regulates the strength of the regularization. RLC controls the complexity of the model by penalizing large gradients, which helps prevent overfitting and promotes better generalization.

\subsection{Mathematical Intuition for Improved Mode Coverage}

To illustrate how the HingeRLC-GAN mitigates mode collapse, we provide a theoretical explanation. The combination of Hinge Loss and Regularized Loss Control (RLC) fosters diverse mode coverage by penalizing the Discriminator for overly confident predictions, which encourages the Generator to explore a broader range of the data distribution.

\subsubsection{Theoretical Framework}

Consider the Hinge Loss function for the Discriminator \( D \):

\[
\mathcal{L}_D = \mathbb{E}_{x \sim p_{\text{data}}} \left[ \max(0, 1 - D(x)) \right] + \mathbb{E}_{z \sim p_z} \left[ \max(0, 1 + D(G(z))) \right]
\]

The gradient of this loss with respect to the Discriminator's parameters \( \theta_D \) is:

\[
\nabla_{\theta_D} \mathcal{L}_D = \mathbb{E}_{x \sim p_{\text{data}}} \left[ \mathbf{1}_{D(x) < 1} \cdot \nabla_{\theta_D} (-D(x)) \right] + \mathbb{E}_{z \sim p_z} \left[ \mathbf{1}_{D(G(z)) > -1} \cdot \nabla_{\theta_D} D(G(z)) \right]
\]

where \( \mathbf{1} \) is the indicator function. The Generator \( G \) minimizes the Hinge Loss:

\[
\mathcal{L}_G = -\mathbb{E}_{z \sim p_z} \left[ D(G(z)) \right]
\]

The gradient of the Generator's loss with respect to its parameters \( \theta_G \) is:

\[
\nabla_{\theta_G} \mathcal{L}_G = -\mathbb{E}_{z \sim p_z} \left[ \nabla_{\theta_G} D(G(z)) \right]
\]

The Regularized Loss Control (RLC) term is introduced to the Hinge Loss to form the total loss for the Discriminator:

\[
\mathcal{L}_D^{\text{RLC}} = \mathcal{L}_D + \lambda \sum_{i} \left( \frac{\partial \mathcal{L}_D}{\partial \theta_{D,i}} \right)^2
\]

This regularization term penalizes large gradients by adding the squared norms of the gradients to the loss function. It discourages the Discriminator from making overly confident predictions, which in turn compels the Generator to explore a more diverse set of data samples.

\subsubsection{Prevention of Mode Collapse}

The integration of Hinge Loss and RLC in the HingeRLC-GAN plays a crucial role in preventing mode collapse. By discouraging the Discriminator from being too confident and smoothing its decision boundaries, the RLC term forces the Generator to explore a wider variety of data modes. This reduces the likelihood of mode collapse, where the Generator might otherwise focus on generating a limited set of samples.


\section{Experimental Analysis}
\label{sec:Results}

The Frechet Inception Distance (FID) \cite{heusel2017gans} is the most commonly used metric for evaluating GAN performance. FID measures the difference between the distributions of features extracted from real and generated images using the InceptionV3 model. This metric provides a more comprehensive evaluation compared to the Inception Score, which only assesses the quality of generated images based on their own features.

FID evaluates both the mean and variance of the feature distributions from real and generated images. A lower FID indicates that the generated images are of higher quality and have better diversity, resembling the real images from the dataset, such as CIFAR-10. While FID assumes Gaussian distributions for the features, it can still be biased for smaller datasets like CIFAR-10 due to the limited sample size.

Mathematically, FID is calculated as:

\[
\text{FID} = \left\| \mu_r - \mu_g \right\|^2 + \text{Tr} \left( \Sigma_r + \Sigma_g - 2 \left( \Sigma_r \Sigma_g \right)^{1/2} \right)
\]

where, \(\mu_r\) and \(\mu_g\) are the means of the feature vectors for real and generated images, respectively. \(\Sigma_r\) and \(\Sigma_g\) are the covariance matrices of the feature vectors for real and generated images, respectively. \(\text{Tr}\) denotes the trace of a matrix.

This formula measures the distance between two multivariate Gaussians defined by their mean vectors and covariance matrices, providing a quantitative measure of how similar the generated images are to the real ones.

The Kernel Inception Distance (KID) is a metric for evaluating GAN-generated images. Unlike Frechet Inception Distance (FID), which assumes Gaussian distributions, KID uses the squared Maximum Mean Discrepancy (MMD) to measure the distance between feature distributions of real and generated images.

  \[
  \text{KID} = \frac{1}{2} \left( \text{MMD}^2(p_r, p_g) + \text{MMD}^2(p_g, p_r) \right)
  \]

  where MMD is computed using a kernel function.

\subsection{Comparison of GAN Architectures}

In Table \ref{tab:tab1}, we compare different GAN architectures using the FID score and KID Score. Each architecture varies in the network used for the generator and discriminator.

\begin{table*}[h]
\caption{\textbf{Comparison to GAN Architecture.}
We report the average FID (↓) scores and average KID (↓) scores on the CIFAR datasets.}
\label{tab:tab1}
\small
\begin{tabular}{|p{0.25\linewidth}|p{0.28\linewidth}|p{0.30\linewidth}|c|c|}
\hline
Architecture & Generator & Discriminator & FID↓ & KID↓ \\ \hline
Dense + VGG & Dense network + BCE & VGG + MinMax & 125 & 0.01\\ \hline
MobileNet & MobileNet + BCE & MobileNet + MinMax & 112 & 0.01 \\ \hline
EfficientNet & EfficientNet + BCE & EfficientNet + MinMax & 97 & 0.01 \\ \hline
ResNet & ResNet blocks + BCE & ResNet blocks + MinMax & \textbf{90} & \textbf{0.002}\\ \hline
\end{tabular}
\end{table*}

The Dense + VGG architecture shows the highest FID score of 125, indicating the poorest performance among the architectures tested. MobileNet improves the FID score to 112. EfficientNet further reduces the FID score to 97, showing better image generation quality. The best performance is observed with the ResNet architecture, achieving an FID score of 90, demonstrating its effectiveness in generating high-quality images.

\subsection{Comparison of Loss Functions}

Table \ref{tab:tab2} compares different GAN loss functions, all using the ResNet architecture for both the generator and discriminator.

\begin{table*}[h]
\caption{\textbf{Comparison to GAN Loss Functions.}
We report the average FID (↓) scores and average KID (↓) scores on the CIFAR datasets.}
\label{tab:tab2}
\small
\begin{tabular}{|p{0.25\linewidth}|p{0.28\linewidth}|p{0.30\linewidth}|c|c|}
\hline
Model & Generator & Discriminator & FID↓ & KID↓ \\ \hline
ResNet (baseline) & ResNet blocks + BCE & ResNet blocks + MinMax & 90 & 0.002 \\ \hline
WGAN-GP with ResNet & ResNet blocks + BCE & ResNet blocks + Wasserstein Loss& 35 & 0.001\\ \hline
lsGAN with ResNet & ResNet blocks + BCE & ResNet blocks + lsGAN & 35 & 0.001\\ \hline
lsGAN with ResNet & ResNet blocks + lsGAN & ResNet blocks + lsGAN & 29 & 0.001\\ \hline
Hinge Loss with ResNet & ResNet blocks + BCE & ResNet blocks + Hinge Loss & \textbf{25} & \textbf{0.001} \\ \hline
\end{tabular}
\end{table*}

The baseline ResNet model with BCE and MinMax loss functions yields an FID score of 90. Using Wasserstein Loss (WGAN-GP) with ResNet significantly improves the FID score to 35. The least squares GAN (lsGAN) with ResNet achieves a similar FID score of 35 with BCE for the generator. When lsGAN is used for both generator and discriminator, the FID improves to 29. The best performance is achieved with Hinge Loss, bringing the FID score down to 25.

\subsection{Comparison of Regularization Methods}

Table \ref{tab:tab3} explores the impact of various regularization methods on the FID scores for the ResNet architecture with Hinge Loss.

\begin{table*}[h]
\caption{\textbf{Comparison to GAN Regularization Methods.}
We report the average FID (↓) scores on the CIFAR datasets.}
\label{tab:tab3}
\small
\begin{tabular}{|p{0.40\linewidth}|p{0.30\linewidth}|p{0.15\linewidth}|p{0.12\linewidth}|}
\hline
Model & Regularization Methods & FID↓ & KID↓ \\ \hline
ResNet with Hinge Loss & No regularization & 25 & 0.001\\ \hline
ResNet with Hinge Loss & + Noise & 25 & 0.001\\ \hline
ResNet with Hinge Loss & + CR & 19 & 0.001\\ \hline
ResNet with Hinge Loss & + GP-0 & 26 & 0.001\\ \hline
ResNet with Hinge Loss (Ours) & + RLC & \textbf{18} & \textbf{0.001} \\ \hline
\end{tabular}
\end{table*}

Without any regularization, the ResNet model with Hinge Loss achieves an FID score of 25. Adding noise does not change the FID score. Contrastive Regularization (CR) improves the FID to 19, while the Gradient Penalty (GP-0) slightly worsens the FID to 26. Our proposed Regularized Loss Control (RLC) method yields the best FID score of 18.

\subsection{Comparison of GAN Models}

Table \ref{tab:tab4} provides a comparison of different GAN models, highlighting the effectiveness of our HingeRLC-GAN model.

\begin{table*}[h]
\caption{\textbf{Comparison of GAN Models.}
We report the average FID (↓) scores and Inception Score (↑) on the CIFAR datasets.}
\label{tab:tab4}
\small
\begin{tabular}{|p{0.40\linewidth}|p{0.20\linewidth}|p{0.20\linewidth}|}
\hline
Model & FID↓ &Inception Score↑\\ \hline
MGO-GAN & 198 & 6.130\\ \hline
DROPOUT-GAN & 66 & -\\ \hline
DCGAN & 53 & 6.47\\ \hline
LSGAN & 56 & 6.32 \\ \hline
DRAGAN & 52 & 6.44\\ \hline
DFM & 52 & 6.58\\ \hline
\textbf{HingeRLC-GAN (Ours) }& \textbf{18} & \textbf{6.89} \\ \hline
\end{tabular}
\end{table*}

The MGO-GAN model shows the highest FID score of 198, indicating the poorest performance. DROPOUT-GAN significantly improves the FID score to 66. DCGAN and LSGAN achieve similar FID scores of 53 and 56, respectively. DRAGAN and DFM further improve the FID to 52. Our proposed HingeRLC-GAN achieves the best FID score of 18, demonstrating superior performance in generating high-quality and diverse images.

\subsection{Mode Capture Analysis}

We first present a t-SNE visualization of the CIFAR-10 dataset images, illustrating the clustering of different classes.

\begin{figure}[ht]
\centering
\includegraphics[width=0.8\textwidth]{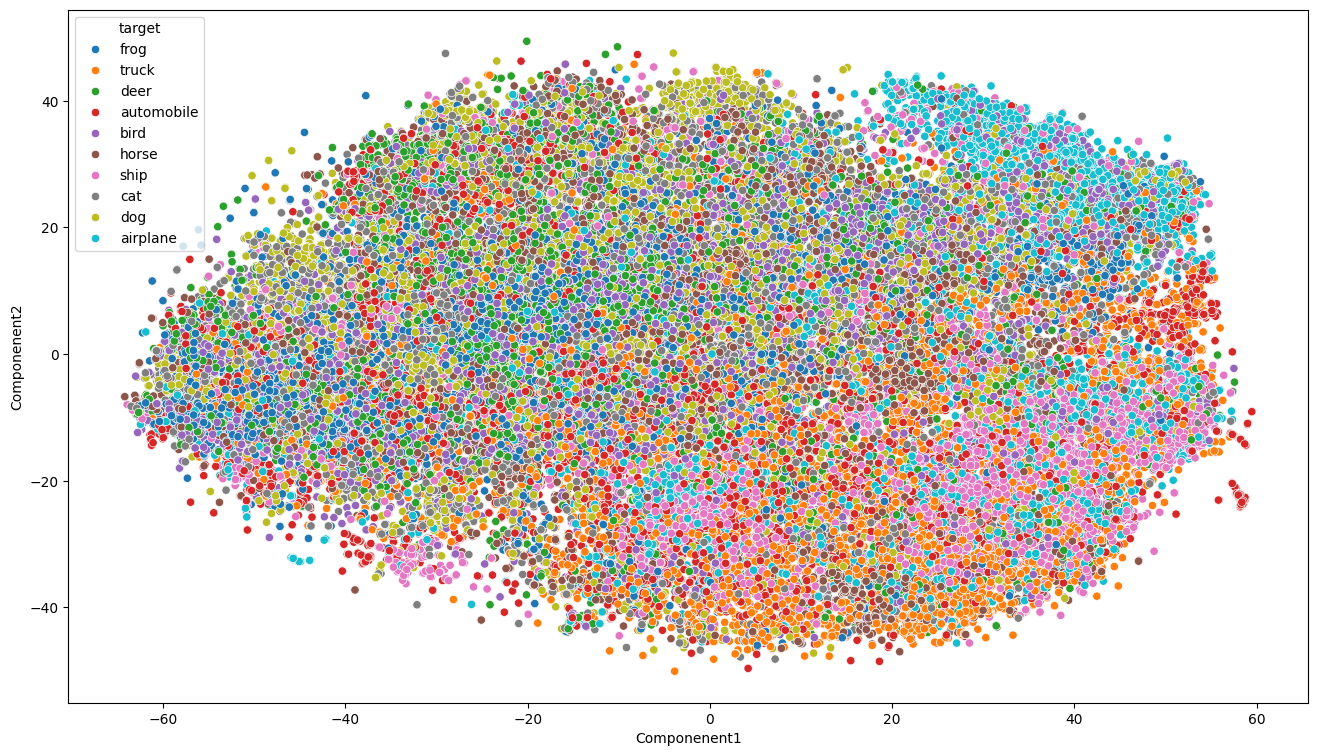}
\caption{t-SNE visualization of the CIFAR-10 dataset images.}
\label{fig:tsne_cifar10}
\end{figure}

In the visualization:
\begin{itemize}
    \item Airplanes are distinctly clustered in the top left corner, indicating clear separability from other classes.
    \item Frogs and cats show significant overlap with other categories and are dispersed across the visualization space.
    \item Automobiles are also spread out, suggesting intra-class variability.
    \item Trucks and ships, while more distinct from other classes, show a degree of overlap between them, located in the bottom left corner.
\end{itemize}

Next, we compare the t-SNE visualizations of images generated by DROPOUT-GAN and our HingeRLC-GAN, demonstrating a 30\% improvement in mode capture with our model.

\begin{figure*}[ht]
    \centering
    \includegraphics[width=0.48\linewidth]{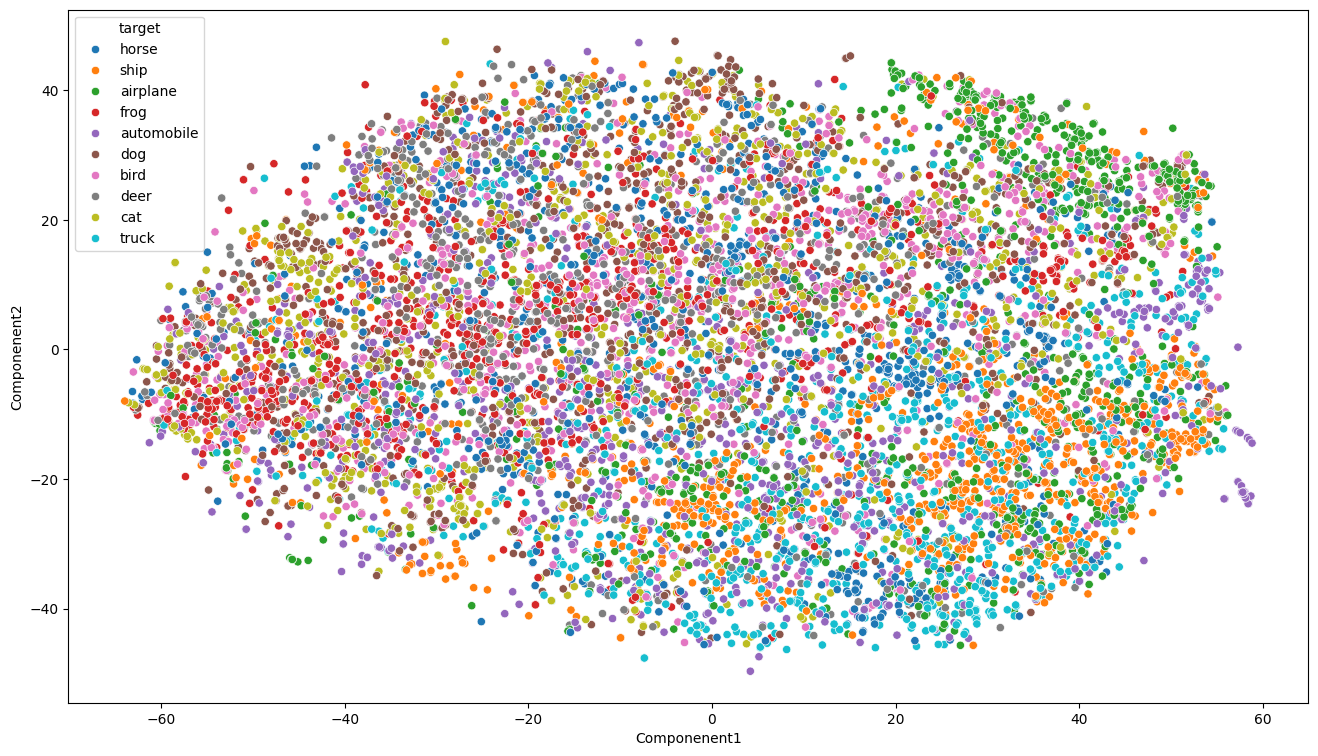}
    \includegraphics[width=0.48\linewidth]{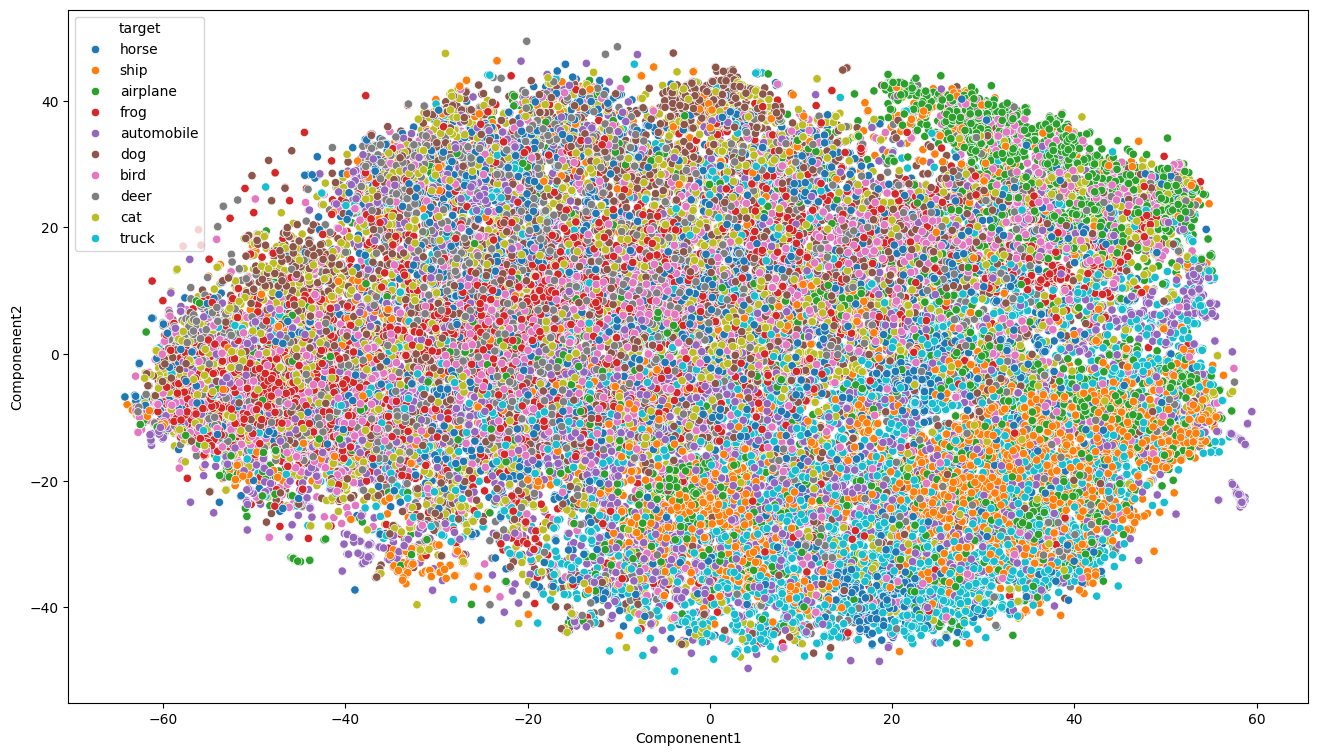}
    \caption{t-SNE Visualizations: (left) DROPOUT-GAN, (right) HingeRLC-GAN. Mode coverage is 30\% better then DROPOUT-GAN}
    \label{fig:tsne_comparison}
\end{figure*}

\subsection{Evaluation of HingeRLC-GAN}

\begin{figure*}[ht]
    \centering
    \includegraphics[width=0.8\textwidth]{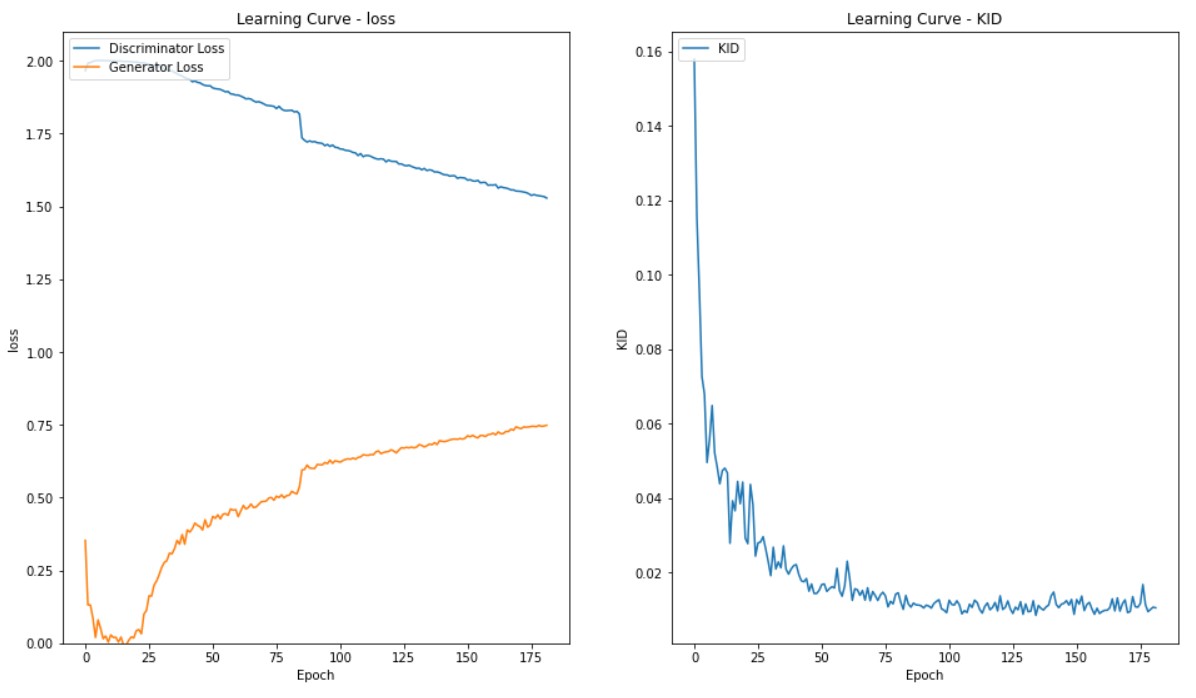}
    \caption{Generator and Discriminator Loss, Learning Curve for HingeRLC-GAN}
    \label{fig:hingeralcgan}
\end{figure*}

The HingeRLC-GAN evaluation yielded several significant results. Despite slower convergence, the model successfully produced high-quality and diverse images without mode collapse. The FID and KID metrics were recorded as 18 and 0.001, respectively. The training process showed gradual fluctuations in both discriminator and generator losses, indicating stable and balanced training dynamics. The KID metric showed a steady decline until about 60 epochs, after which it plateaued. The learning rate was reduced to 0.00005 after 80 epochs to maintain equilibrium between discriminator and generator losses as shown in Figure \ref{fig:hingeralcgan}.

\subsection{Generated Images}

\begin{figure*}[ht]
    \centering
    \includegraphics[width=0.9\linewidth]{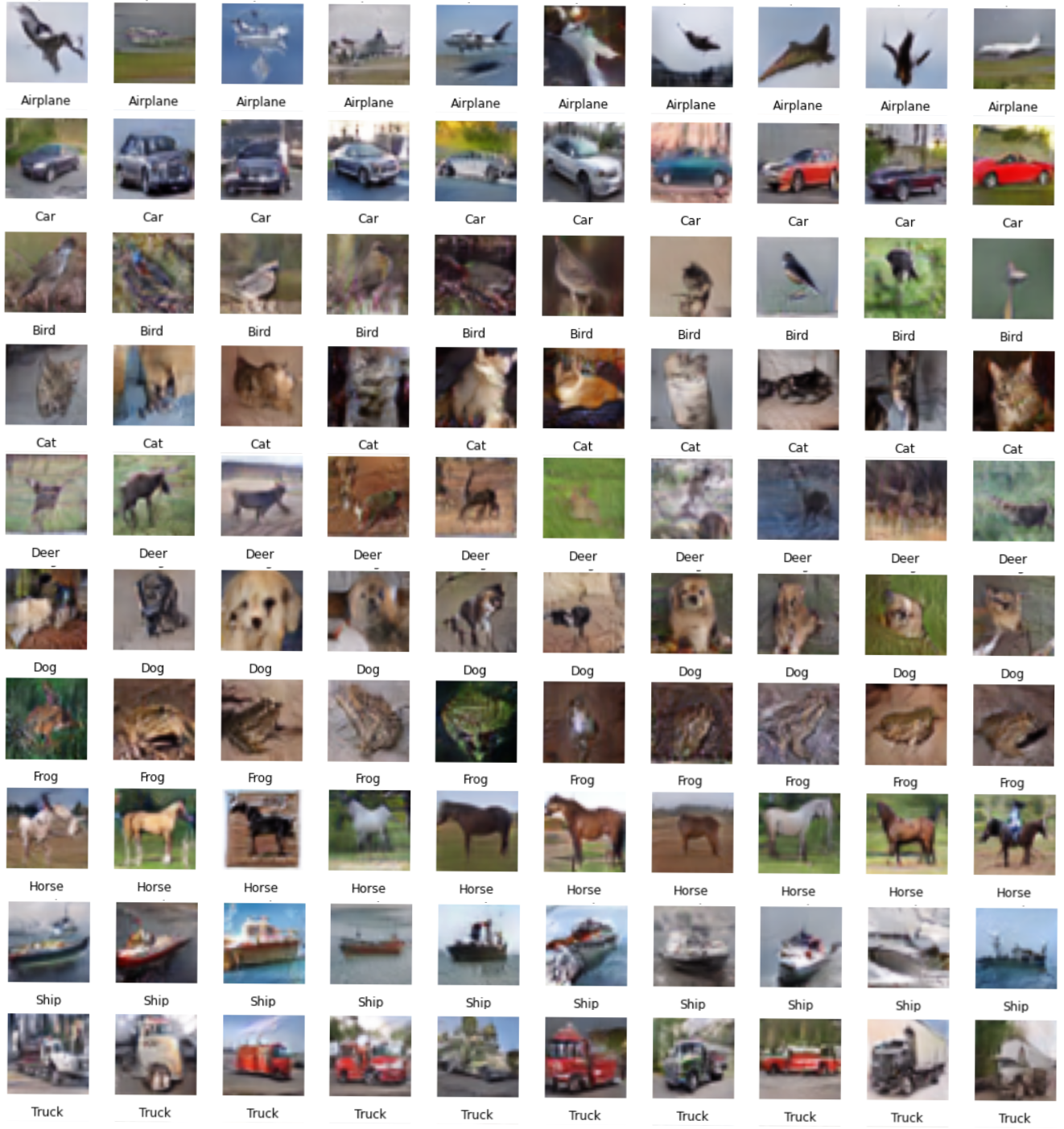}
    \caption{Sample Images Generated by HingeRLC-GAN}
    \label{fig:HingeRLC_gan_images}
\end{figure*}

We present 10 images representing 10 different classes generated by the HingeRLC-GAN. Notably, vehicles (especially ships), birds, horses, deers, and dogs appeared realistic. Some anomalies were observed in frog and cat images, likely due to the intrinsic diversity within these classes in the CIFAR-10 dataset as shown in Figure \ref{fig:HingeRLC_gan_images}. The anomalies in frog and cat images may be due to the intricate variations found within these specific classes in the CIFAR-10 dataset. Despite these anomalies, the HingeRLC-GAN's overall realism across various categories proves its suitability for a wide range of image generation tasks.

\section{Conclusion}
\label{sec:Conclusion}

In this paper, we introduced  HingeRLC GAN, a novel variant that addresses mode collapse by integrating Hinge Loss with Regularized Loss Control (RLC). Our experiments demonstrate that this approach significantly enhances both the diversity and quality of generated images. Through extensive evaluation of various GAN architectures, we found ResNet to be the most effective baseline, and HingeRLC GAN consistently outperformed traditional loss functions like Wasserstein and Least Squares, as well as other regularization techniques, achieving the lowest FID scores. The HingeRLC GAN surpassed state-of-the-art models, including MGO-GAN, DROPOUT-GAN, DCGAN, LSGAN, DRAGAN, and DFM, proving its superiority in generating diverse and high-fidelity images. Overall, HingeRLC GAN provides a robust solution to mode collapse, enhancing training stability and output quality, with future research focusing on its application across diverse datasets and tasks in generative modeling and computer vision.

 \end{document}